\documentclass[10pt,twocolumn,letterpaper]{article}

\usepackage{cvpr}              

\usepackage[accsupp]{axessibility}
\usepackage{graphicx}
\usepackage{amsmath}
\usepackage{amssymb}
\usepackage{booktabs}
\usepackage{times}
\usepackage{epsfig}
\usepackage{graphicx}
\usepackage{tabularx}
\usepackage{arydshln}
\usepackage{bm}
\usepackage{nicefrac}
\usepackage{fixltx2e}
\usepackage{multirow}
\usepackage{mathtools}
\usepackage{algorithm}
\usepackage{algpseudocode}
\makeatletter
\@namedef{ver@everyshi.sty}{}
\makeatother
\usepackage{tikz}
\usepackage{tikz}
\usetikzlibrary{arrows,shapes,calc,matrix,fit,backgrounds}
\usepackage{pgfplots}
\usepackage{pgfplotstable}
\pgfplotsset{compat=1.9}
\usepgfplotslibrary{fillbetween}

\usepackage{xstring}

\usepgfplotslibrary{external}

\IfBeginWith*{\jobname}{fig/extern/}{\finalcopy}{}

\tikzset{every mark/.append style={solid}}
\pgfplotsset{
	grid=both, width=\columnwidth, try min ticks=5,
	every axis/.append style={font=\scriptsize},
	every axis plot/.append style={thick,mark=none,mark size=1.2,tension=0.18},
	legend cell align=left, legend style={fill opacity=0.8},
}

\pgfplotsset{
	dash/.style={mark=o,dashed,opacity=0.7},
	dott/.style={mark=o,dotted,opacity=0.7},
}

\newcommand{\tablestyle}[2]{\setlength{\tabcolsep}{#1}\renewcommand{\arraystretch}{#2}\centering\footnotesize}

\usepackage[pagebackref,breaklinks,    colorlinks]{hyperref}

\usepackage[capitalize]{cleveref}
\crefname{section}{Sec.}{Secs.}
\Crefname{section}{Section}{Sections}
\Crefname{table}{Table}{Tables}
\crefname{table}{Tab.}{Tabs.}

\def\cvprPaperID{4801} 
\def\confName{CVPR}
\def\confYear{2022}

\renewcommand{\paragraph}[1]{\noindent\textbf{#1}}

\begin{document}

\title{Recall@k Surrogate Loss with Large Batches and Similarity Mixup}

\author{Yash Patel \quad Giorgos Tolias \quad Ji{\v{r}}{\'i} Matas \\
Visual Recognition Group, Czech Technical University in Prague \\
{\tt\small \{patelyas,toliageo,matas\}@fel.cvut.cz}}
\maketitle
\newcommand{\nn}[1]{\ensuremath{\text{NN}_{#1}}\xspace}
\def\l1{\ensuremath{\ell_1}\xspace}
\def\l2{\ensuremath{\ell_2}\xspace}

\def\roxf{$\mathcal{R}$Oxford\xspace}
\def\rox{$\mathcal{R}$Oxf\xspace}
\def\ro{$\mathcal{R}$O\xspace}
\def\rpar{$\mathcal{R}$Paris\xspace}
\def\rpa{$\mathcal{R}$Par\xspace}
\def\rp{$\mathcal{R}$P\xspace}
\def\rdis{$\mathcal{R}$1M\xspace}

\newcommand\resnet[3]{\ensuremath{\prescript{#2}{}{\mathtt{R}}{#1}_{\scriptscriptstyle #3}}\xspace}

\newcommand*\OK{\ding{51}}

\newenvironment{narrow}[1][1pt]
	{\setlength{\tabcolsep}{#1}}
	{\setlength{\tabcolsep}{6pt}}

\newcommand{\alert}[1]{{\color{red}{#1}}}
\newcommand{\gio}[1]{{\color{blue}{#1}}}
\newcommand{\replace}[2]{{\color{gray}{#1}}{\color{red}{#2}}}

\newcommand{\comment} [1]{{\color{orange} \Comment     #1}} 


\newcommand{\head}[1]{{\smallskip\noindent\bf #1}}
\newcommand{\equ}[1]{(\ref{equ:#1})\xspace}

\newcommand{\red}[1]{{\color{red}{#1}}}
\newcommand{\blue}[1]{{\color{blue}{#1}}}
\newcommand{\green}[1]{{\color{green}{#1}}}
\newcommand{\gray}[1]{{\color{gray}{#1}}}


\newcommand{\tran}{^\top}
\newcommand{\mtran}{^{-\top}}
\newcommand{\zcol}{\mathbf{0}}
\newcommand{\zrow}{\zcol\tran}

\newcommand{\ind}{\mathds{1}}
\newcommand{\expect}{\mathbb{E}}
\newcommand{\nat}{\mathbb{N}}
\newcommand{\zahl}{\mathbb{Z}}
\newcommand{\real}{\mathbb{R}}
\newcommand{\proj}{\mathbb{P}}
\newcommand{\prob}{\mathbf{Pr}}

\newcommand{\mif}{\textrm{if }}
\newcommand{\other}{\textrm{otherwise}}
\newcommand{\minimize}{\textrm{minimize }}
\newcommand{\maximize}{\textrm{maximize }}

\newcommand{\id}{\operatorname{id}}
\newcommand{\const}{\operatorname{const}}
\newcommand{\sgn}{\operatorname{sgn}}
\newcommand{\var}{\operatorname{Var}}
\newcommand{\mean}{\operatorname{mean}}
\newcommand{\trace}{\operatorname{tr}}
\newcommand{\diag}{\operatorname{diag}}
\newcommand{\vect}{\operatorname{vec}}
\newcommand{\cov}{\operatorname{cov}}

\newcommand{\softmax}{\operatorname{softmax}}
\newcommand{\clip}{\operatorname{clip}}

\newcommand{\defn}{\mathrel{:=}}
\newcommand{\peq}{\mathrel{+\!=}}
\newcommand{\meq}{\mathrel{-\!=}}

\newcommand{\floor}[1]{\left\lfloor{#1}\right\rfloor}
\newcommand{\ceil}[1]{\left\lceil{#1}\right\rceil}
\newcommand{\inner}[1]{\left\langle{#1}\right\rangle}
\newcommand{\norm}[1]{\left\|{#1}\right\|}
\newcommand{\frob}[1]{\norm{#1}_F}
\newcommand{\card}[1]{\left|{#1}\right|\xspace}
\newcommand{\diff}{\mathrm{d}}
\newcommand{\der}[3][]{\frac{d^{#1}#2}{d#3^{#1}}}
\newcommand{\pder}[3][]{\frac{\partial^{#1}{#2}}{\partial{#3^{#1}}}}
\newcommand{\ipder}[3][]{\partial^{#1}{#2}/\partial{#3^{#1}}}
\newcommand{\dder}[3]{\frac{\partial^2{#1}}{\partial{#2}\partial{#3}}}

\newcommand{\wb}[1]{\overline{#1}}
\newcommand{\wt}[1]{\widetilde{#1}}

\def\nsp{\hspace{-3pt}}
\def\zsp{\hspace{0pt}}
\def\xssp{\hspace{1pt}}
\def\ssp{\hspace{3pt}}
\def\msp{\hspace{6pt}}
\def\lsp{\hspace{12pt}}
\def\xlsp{\hspace{20pt}}

\newcommand{\cA}{\mathcal{A}}
\newcommand{\cB}{\mathcal{B}}
\newcommand{\cC}{\mathcal{C}}
\newcommand{\cD}{\mathcal{D}}
\newcommand{\cE}{\mathcal{E}}
\newcommand{\cF}{\mathcal{F}}
\newcommand{\cG}{\mathcal{G}}
\newcommand{\cH}{\mathcal{H}}
\newcommand{\cI}{\mathcal{I}}
\newcommand{\cJ}{\mathcal{J}}
\newcommand{\cK}{\mathcal{K}}
\newcommand{\cL}{\mathcal{L}}
\newcommand{\cM}{\mathcal{M}}
\newcommand{\cN}{\mathcal{N}}
\newcommand{\cO}{\mathcal{O}}
\newcommand{\cP}{\mathcal{P}}
\newcommand{\cQ}{\mathcal{Q}}
\newcommand{\cR}{\mathcal{R}}
\newcommand{\cS}{\mathcal{S}}
\newcommand{\cT}{\mathcal{T}}
\newcommand{\cU}{\mathcal{U}}
\newcommand{\cV}{\mathcal{V}}
\newcommand{\cW}{\mathcal{W}}
\newcommand{\cX}{\mathcal{X}}
\newcommand{\cY}{\mathcal{Y}}
\newcommand{\cZ}{\mathcal{Z}}

\newcommand{\vA}{\mathbf{A}}
\newcommand{\vB}{\mathbf{B}}
\newcommand{\vC}{\mathbf{C}}
\newcommand{\vD}{\mathbf{D}}
\newcommand{\vE}{\mathbf{E}}
\newcommand{\vF}{\mathbf{F}}
\newcommand{\vG}{\mathbf{G}}
\newcommand{\vH}{\mathbf{H}}
\newcommand{\vI}{\mathbf{I}}
\newcommand{\vJ}{\mathbf{J}}
\newcommand{\vK}{\mathbf{K}}
\newcommand{\vL}{\mathbf{L}}
\newcommand{\vM}{\mathbf{M}}
\newcommand{\vN}{\mathbf{N}}
\newcommand{\vO}{\mathbf{O}}
\newcommand{\vP}{\mathbf{P}}
\newcommand{\vQ}{\mathbf{Q}}
\newcommand{\vR}{\mathbf{R}}
\newcommand{\vS}{\mathbf{S}}
\newcommand{\vT}{\mathbf{T}}
\newcommand{\vU}{\mathbf{U}}
\newcommand{\vV}{\mathbf{V}}
\newcommand{\vW}{\mathbf{W}}
\newcommand{\vX}{\mathbf{X}}
\newcommand{\vY}{\mathbf{Y}}
\newcommand{\vZ}{\mathbf{Z}}

\newcommand{\va}{\mathbf{a}}
\newcommand{\vb}{\mathbf{b}}
\newcommand{\vc}{\mathbf{c}}
\newcommand{\vd}{\mathbf{d}}
\newcommand{\ve}{\mathbf{e}}
\newcommand{\vf}{\mathbf{f}}
\newcommand{\vg}{\mathbf{g}}
\newcommand{\vh}{\mathbf{h}}
\newcommand{\vi}{\mathbf{i}}
\newcommand{\vj}{\mathbf{j}}
\newcommand{\vk}{\mathbf{k}}
\newcommand{\vl}{\mathbf{l}}
\newcommand{\vm}{\mathbf{m}}
\newcommand{\vn}{\mathbf{n}}
\newcommand{\vo}{\mathbf{o}}
\newcommand{\vp}{\mathbf{p}}
\newcommand{\vq}{\mathbf{q}}
\newcommand{\vr}{\mathbf{r}}
\newcommand{\Vs}{\mathbf{s}}
\newcommand{\vt}{\mathbf{t}}
\newcommand{\vu}{\mathbf{u}}
\newcommand{\vv}{\mathbf{v}}
\newcommand{\vw}{\mathbf{w}}
\newcommand{\vx}{\mathbf{x}}
\newcommand{\vy}{\mathbf{y}}
\newcommand{\vz}{\mathbf{z}}

\newcommand{\vone}{\mathbf{1}}
\newcommand{\vzero}{\mathbf{0}}

\newcommand{\valpha}{{\boldsymbol{\alpha}}}
\newcommand{\vbeta}{{\boldsymbol{\beta}}}
\newcommand{\vgamma}{{\boldsymbol{\gamma}}}
\newcommand{\vdelta}{{\boldsymbol{\delta}}}
\newcommand{\vepsilon}{{\boldsymbol{\epsilon}}}
\newcommand{\vzeta}{{\boldsymbol{\zeta}}}
\newcommand{\veta}{{\boldsymbol{\eta}}}
\newcommand{\vtheta}{{\boldsymbol{\theta}}}
\newcommand{\viota}{{\boldsymbol{\iota}}}
\newcommand{\vkappa}{{\boldsymbol{\kappa}}}
\newcommand{\vlambda}{{\boldsymbol{\lambda}}}
\newcommand{\vmu}{{\boldsymbol{\mu}}}
\newcommand{\vnu}{{\boldsymbol{\nu}}}
\newcommand{\vxi}{{\boldsymbol{\xi}}}
\newcommand{\vomikron}{{\boldsymbol{\omikron}}}
\newcommand{\vpi}{{\boldsymbol{\pi}}}
\newcommand{\vrho}{{\boldsymbol{\rho}}}
\newcommand{\vsigma}{{\boldsymbol{\sigma}}}
\newcommand{\vtau}{{\boldsymbol{\tau}}}
\newcommand{\vupsilon}{{\boldsymbol{\upsilon}}}
\newcommand{\vphi}{{\boldsymbol{\phi}}}
\newcommand{\vchi}{{\boldsymbol{\chi}}}
\newcommand{\vpsi}{{\boldsymbol{\psi}}}
\newcommand{\vomega}{{\boldsymbol{\omega}}}

\newcommand{\rLambda}{\mathrm{\Lambda}}
\newcommand{\rSigma}{\mathrm{\Sigma}}

\makeatletter
\DeclareRobustCommand\onedot{\futurelet\@let@token\@onedot}
\def\@onedot{\ifx\@let@token.\else.\null\fi\xspace}
\def\eg{\emph{e.g}\onedot} \def\Eg{\emph{E.g}\onedot}
\def\ie{\emph{i.e}\onedot} \def\Ie{\emph{I.e}\onedot}
\def\cf{\emph{cf}\onedot} \def\Cf{\emph{C.f}\onedot}
\def\etc{\emph{etc}\onedot} \def\vs{\emph{vs}\onedot}
\def\wrt{w.r.t\onedot} \def\dof{d.o.f\onedot}
\def\etal{\emph{et al}\onedot}
\makeatother

\begin{abstract}
This work focuses on learning deep visual representation models for retrieval by exploring the interplay between a new loss function, the batch size, and a new regularization approach. Direct optimization, by gradient descent, of an evaluation metric, is not possible when it is non-differentiable, which is the case for recall in retrieval. A differentiable surrogate loss for the recall is proposed in this work. Using an implementation that sidesteps the hardware constraints of the GPU memory, the method trains with a very large batch size, which is essential for metrics computed on the entire retrieval database. It is assisted by an efficient mixup regularization approach that operates on pairwise scalar similarities and virtually increases the batch size further. The suggested method achieves state-of-the-art performance in several image retrieval benchmarks when used for deep metric learning. For instance-level recognition, the method outperforms similar approaches that train using an approximation of average precision. 
\end{abstract}

\vspace{10pt}
\section{Introduction}
\label{sec:introduction}

Minimization of a loss that is a function of the test-time evaluation metric has shown to be beneficial in deep learning for numerous computer vision and natural language processing tasks. Examples include intersection-over-union as a loss that boosts performance for object detection~\cite{yjw+16,rig+19} and semantic segmentation~\cite{nsb+18}, and structural similarity~\cite{mae+18}, peak signal-to-noise ratio~\cite{bms+18} and perceptual~\cite{pam+21} as reconstruction losses for image compression that give better results according to the respective evaluation metrics. 

\begin{figure}[t]
\vspace{-10pt}
\begin{center}
\includegraphics[width=0.45\textwidth]{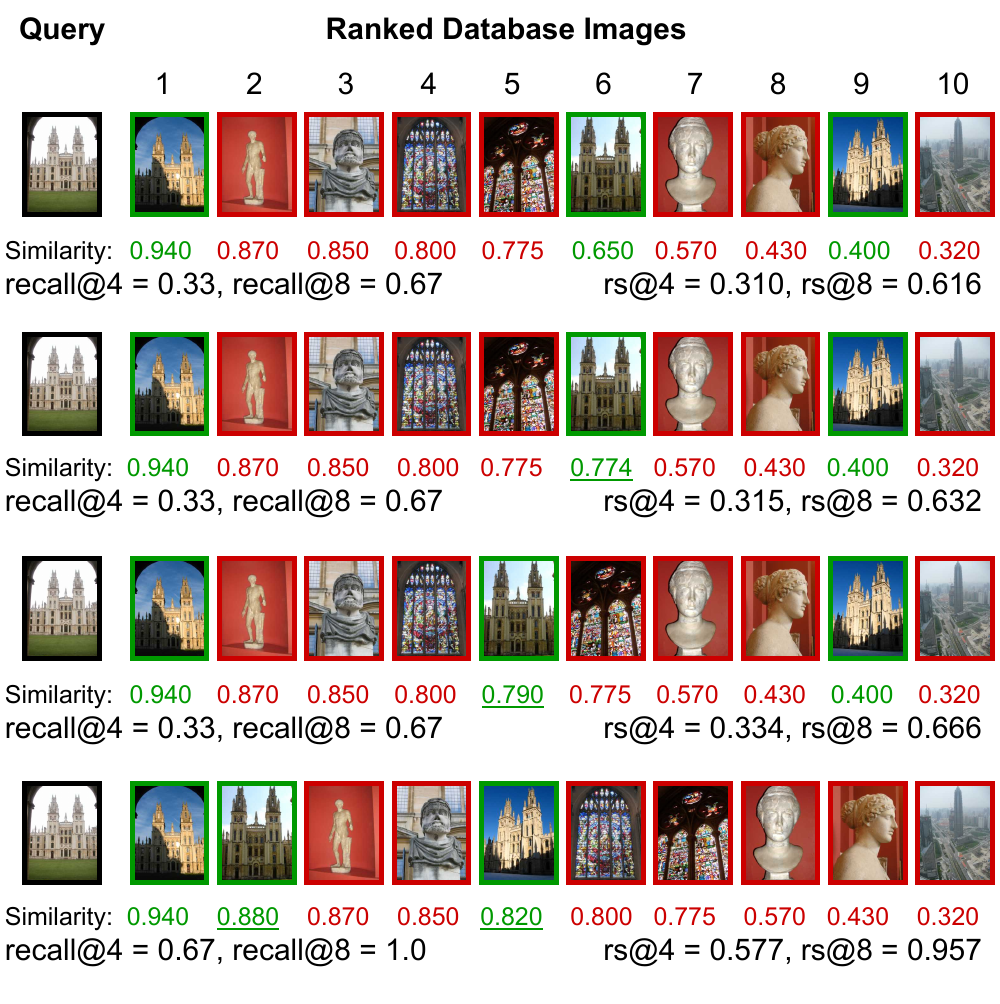}
\vspace{-1.2em}
\caption{A comparison between recall@k and rs@k, the proposed differentiable recall@k surrogate. Examples show a query, the ranked database images sorted according to the similarity and the corresponding values for recall@k and rs@k and their dependence on similarity score change. Note that the values of recall@k and rs@k are close. Changes to similarity and ranking in some cases may not affect the original recall@k but can affect the surrogate, with the latter having a more significant impact than the former. Similarity values of all negatives are fixed for ease of understanding. The similarity values of the positives that were changed in rows 2, 3 and 4 are \underline{underlined}.
\label{fig:teaser}}
\end{center}
\vspace{-2em}
\end{figure}

Training deep networks via gradient descent on the evaluation metric is not possible when the metric is non-differentiable. Deep learning methods resort to a proxy loss, a differentiable function, as a workaround, which empirically leads to a reasonable performance but may not align well with the evaluation metric. Examples exist in object detection~\cite{yjw+16}, scene text recognition~\cite{phm+20,pm+21}, machine translation~\cite{bbx+17} and image retrieval~\cite{bxk+20,pgr+19}.

This paper deals with the training of image retrieval posed  as deep metric learning and Euclidean search in the learned image embedding space. It is the task of ranking all database examples according to the relevance to a query, which is of vital importance for many applications. The standard evaluation metrics are precision and recall in the top retrieved results and the mean Average Precision (mAP).
These metrics are standard in information retrieval, they reflect the quality of the retrieved results and allow for flexibility to focus either on the few top results or the whole ranked list of examples, respectively.
Recall at top-$k$ retrieved results, denoted by \emph{recall@k} in the following, is the primary focus of this work.

The problem related to the optimization of non-differentiable evaluation metrics applies to recall@k as well. 
Estimating the position of positive images in the list of retrieved results and counting how many positives appear inside a short-list of a fixed size involves non-differentiable operations.
Note that methods for training on non-differentiable losses, such as actor-critic~\cite{bbx+17} and learning surrogates~\cite{phm+20} are not directly applicable to recall@k. This is due to the fact that these methods are limited to decomposable functions, where a per-example performance measure is available. Such an attempt is made by Engilberge~\etal~\cite{ecp+19}, where an LSTM learns sorting-based metrics, but is not adapted in consequent work due to slow training. As an alternative, deep metric learning approaches for image retrieval often use ranking proxy losses, termed pairwise losses. In the embedding space, loss functions such as contrastive~\cite{hcl06}, triplet~\cite{skp+15}, and margin~\cite{wms+17} pull the examples from the same class closer to one another and push the examples from a different class away. These losses are hand-crafted to reflect the objectives of the retrieval task and, consequently, the evaluation metric. The loss value depends on the image-to-image similarity for image pairs or triplets and does not take into account the whole ranked list of examples. Changes in the similarity value without any change in the overall ranking alter the loss value indicate that they are not well correlated with ranking~\cite{bxk+20}. Recent methods focus on optimizing Average Precision (AP) and use a surrogate function as a loss~\cite{hls18,rar+19,rmp+20,chx+19,bxk+20}. A surrogate of an evaluation metric is a function that approximates it in a differentiable manner. 

The proposed method attains state-of-the-art results for 4 fine-grained retrieval datasets, namely iNaturalist~\cite{vms+18}, VehicleID~\cite{vms+18}, SOP~\cite{ohb16} and Cars196~\cite{ksd+13}, and 2 instance-level retrieval datasets, namely Revisited Oxford and Paris~\cite{rit+18}. 
This is accomplished by the demonstrated synergy between the three following elements.
First, a new loss that is proposed as a surrogate of an established retrieval evaluation metric, namely recall at top $k$, and is experimentally shown to consistently outperform existing competitors. 
A comparison between the evaluation metric and the proposed loss is shown in Figure \ref{fig:teaser}.
Second, the use of a very large batch size, in the order of several thousand large resolution images on a single GPU. This is inspired by the instance-level retrieval literature \cite{rar+19} and is introduced for the first time in the context of fine-grained categorization. 
In a recent work of verifying prior results in deep metric learning for fine-grained categorization~\cite{mbl20} the batch-size is considered fixed to a single and small value among a large set of comparisons for different losses; in this work we reach batch-sizes that are two orders of magnitude larger than in the work of Musgrave~\etal~\cite{mbl20}.
The third elements is the proposed mixup regularization technique that is computationally efficient and that virtually enlarges the batch. Its efficiency is obtained by operating on the very last stage of similarity estimation, \ie scalar similarities are mixed, while its applicability goes beyond the combination with the proposed loss in this work.
The proposed loss is used for training widely used ResNet architectures~\cite{hzr+16} but also recent vision-transformers (ViT)~\cite{dbk+21}. The superiority of this loss compared to existing losses is demonstrated with both architectures, while with ViT-B/16 top results are achieved at lower throughput than with ResNet.
\vspace{10pt}
\section{Related work}
\label{sec:related_work}

In this section, the related work is reviewed for two different families of deep metric learning approaches regarding the type of loss that is optimized, namely classification losses and pairwise losses. 
Given an embedding network that maps input images to a high dimensional space, in the former, the loss is a function of the embedding and the corresponding category label of a single image, while in the latter, the loss is a function of the distance, or similarity, between two embeddings and the corresponding pairwise label. 
Prior work for mixup~\cite{zcd+17} techniques related to embedding learning is reviewed too. 

\paragraph{Classification losses.} The work of Zhai and Wu~\cite{zw18} supports that the standard classification loss, \ie cross-entropy (CE) loss is a strong approach for deep metric learning. Their finding is supported by the use of layer normalization and class-balanced sampling. In the domain of metric learning for faces, several different classification losses are proposed, such as SphereFace~\cite{lwy+17}, CosFace~\cite{wwz+18} and ArcFace~\cite{dgx+19}, where contributions are in the spirit of large margin classification. Despite the specificity of the domain, such losses are applicable beyond faces. Another variant is the Neighborhood Component Analysis (NCA) loss that is used in the work of Movshovitz-Attias~\etal~\cite{mtl+17}, which is later improved~\cite{tdt20} by temperature-based scaling and faster update of the class prototype vectors, also called proxies in their work. The restriction of a single prototype vector per class is dropped by Qian~\etal~\cite{qss+19} who stores multiple representatives per category. 

Classification losses, in contrast to pairwise losses, perform the optimization independently per image. An exception is the work of Elezi~\etal~\cite{evt+20} where a similarity propagation module captures group interactions within the batch. Then, cross-entropy loss is used, which now comes with significant improvements by taking into account such interactions. This is recently improved~\cite{sel21} by replacing the propagation module with an attention model.
The relation between CE loss and some of the widely used pairwise losses is studied from a mutual information point of view~\cite{brz+20}. CE loss is viewed as approximate bound-optimization for minimizing pairwise losses; CE maximizes mutual information, and so do these pairwise losses, which are reviewed in the following. 

\paragraph{Pairwise losses.} The first pairwise loss introduced for this task is the so-called contrastive loss~\cite{hcl06}, where embeddings of relevant pairs are pushed as close as possible, while those of non-relevant ones are pushed far enough. Since the target task is typically a ranking one, the triplet loss~\cite{skp+15}, a popular and widely used loss, improves that by forming training triplets in the form of anchor, positive and negative examples. The loss is a function of the difference between anchor-to-positive and anchor-to-negative distances and is zero if such a difference is large enough, therefore satisfying the objectives of a ranking task for this triplet. Optimization over all pairs or triplets is not tractable and is observed to be sub-optimal~\cite{wms+17}. As a result, a lot of attention is paid to finding informative pairs and triplets~\cite{mbl20,rms+20,sxj+15,sohn16,lxz+19}, which typically includes heuristics. Several other losses are suggested in the literature~\cite{wms+17,wzw+17,sxj+15} and are added to the long list of hand-designed proxy losses which target to learn embeddings that transfer well to a ranking or a similar task. 

A few cases follow a principled approach for obtaining a loss that is appropriate for ranking tasks. This is the case with the work of Ustinova~\etal~\cite{ul16} where the goal is to minimize the probability that the similarity between embeddings of a non-relevant pair is larger than that of a relevant one. This probability is approximated by the quantization of the range of possible similarities and the histogram loss, which is estimated within a single batch. Their work dispenses with the need for any kind of sampling for mini-batch construction. 
An information-theoretic loss function, called RankMI~\cite{kpd+20}, maximizes the mutual information between the samples within the same semantic class using a neural network. 
Another principled approach focuses on optimizing AP, which is a standard retrieval evaluation metric. A smooth approximation of it is often used in the literature~\cite{rmp+20,hls18,rar+19}, while the work of Brown~\etal~\cite{bxk+20} is the closest to ours. In combination with such AP-based losses, a large batch size is crucial, which meets the limitations set by the hardware. Such limitations are overcome in the work of Revaud~\etal~\cite{rar+19} who uses a batch of $4,000$ high-resolution images.

\paragraph{Embedding mixup.} Manifold mixup~\cite{vlb+19}, which involves mixing~\cite{zcd+17} intermediate representations and labels of two examples, has demonstrated to improve generalizability for supervised learning by encouraging smoother decision boundaries. Such techniques are investigated for embedding learning and image retrieval by mixing the embedding of two examples. Duan~\etal~\cite{dzl+18} uses adversarial training to synthesize additional negative samples from the observed negatives. Kalantidis ~\etal~\cite{ksp+20} synthesize hard-negatives for contrastive self-supervised learning by mixing the embedding of the two hardest negatives and also mixing them with the query itself. Zheng~\etal~\cite{zcl+19} uses a linear interpolation between the embeddings to manipulate the hardness levels. In the work of Gu ~\etal~\cite{gk+20}, two embedding vectors from the same class are used to generate symmetrical synthetic examples and hard-negative mining is performed within the set of original and the synthetic examples. This is further extended to proxy-based losses, where the embedding of examples from different classes and labels is mixed to generate synthetic proxies~\cite{gkk+21}. Linearly interpolating labels entails the risk of generating false negatives if the interpolation factor is close to $0$ or $1$. Such limitations are overcome in the work of Venkataramanan ~\etal~\cite{vpa+21}, which generalizes mixing examples from different classes for pairwise loss functions. The proposed {\em SiMix} approach differs from the aforementioned techniques as it operates on the similarity scores instead of the embedding vectors, does not require training an additional model, making it computationally efficient. Furthermore, unlike the existing mixup techniques, it uses a synthetic sample in the roles of a query, positive and negative example. 
\section{Method}
\label{sec:method}

This section presents the task of image retrieval and the proposed approach for learning image embeddings.

\paragraph{Task.} We are given a query example $q\in \cX$ and a collection of examples $\Omega \subset \cX$, also called database, where $\cX$ is the space of all images. The set of database examples that are positive or negative to the query are denoted by $P_q$ and $N_q$, respectively, with $\Omega=P_q \cup N_q$. Ground-truth information for the positive and negative sets per query is obtained according to discrete class labels per example, \ie if two examples come from the same class, then they are considered positive to each other, otherwise negative. This is the case for all (training or testing) databases used in this work. Terms example and image are used interchangeably in the following text. In image retrieval, all database images are ranked according to similarity to the query $q$, and the goal is to rank positive examples before negative ones.

\paragraph{Deep image embeddings.} 
Image embeddings, otherwise called descriptors, are generated by function $f_{\theta}: \cX \rightarrow \real^d$. In this work, function $f_\theta$ is a deep fully  convolutional neural network or a vision transformer mapping input images of any size or aspect ratio to an $L_{2}$-normalized $d$-dimensional embedding. Embedding for image $x$ is denoted by $\vx = f_\theta(x)$. Parameter set $\theta$ of the network is learned during the training. Similarity between a query $q$ and a database image $x$ is computed by the dot product of the corresponding embeddings and is denoted by $s(q,x) = \vq^\top \vx$, also denoted as $s_{qx}$ for brevity.

\paragraph{Evaluation metric.} Recall@k is one of the standard metrics to evaluate image retrieval methods. For query $q$, it is defined as a ratio of the number of relevant (positive) examples within the top-k ranked examples to the total number of relevant  examples for $q$ given by $|P_q|$.
It is denoted by $R_\Omega^k(q)$ when computed for query $q$ and database $\Omega$ and can be expressed as

\begin{equation}
    R_\Omega^k(q) = \frac{\sum\limits_{x \in P_q} H (k - r_\Omega(q,x))}{|P_q|},
    \label{equ:recall}
\end{equation}
where $r_\Omega(q,x)$ is the rank of example $x$ when all database examples in $\Omega$ are ranked according to similarity to query $q$. Function $H(.)$ is the Heaviside step function, which is equal to 0 for negative values, otherwise equal to 1. The rank of example $x$ is computed by

\begin{equation}
    r_\Omega(q,x) = 1 + \sum\limits_{z \in \Omega, z \neq x} H(s_{qz} - s_{qx}),
    \label{equ:rank}
\end{equation}

Therefore, \equ{recall} can now be expressed as
\begin{equation}
R_\Omega^k(q) = \frac{\sum\limits_{x \in P_q} H (k  - 1 - \sum\limits_{z \in \Omega, z \neq x} H(s_{qz} - s_{qx}) )}{|P_q|}.
\label{equ:recall_full}
\end{equation}

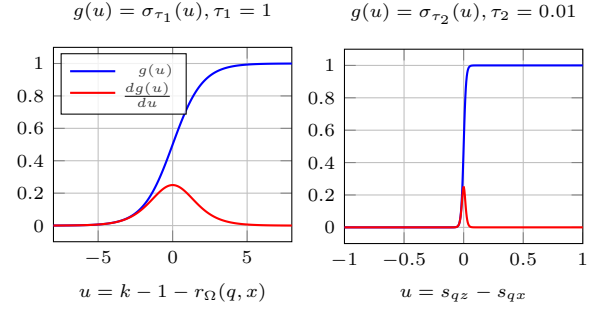
\begin{figure}[t]
\begin{center}
\begin{tikzpicture}[declare function={sigma(\x)=1/(1+exp(-\x));sigmap(\x)=sigma(\x)*(1-sigma(\x));}]
\begin{axis}[%
	width=0.57\linewidth,
	height=0.5\linewidth,
	xlabel={$u = k -1 - r_\Omega(q,x)$},
	title={$g(u)= \sigma_{\tau_1}(u), \tau_1=1$},
	legend cell align={left},
	legend pos=north west,
    legend style={cells={anchor=east}, font =\tiny, fill opacity=0.8, row sep=-2.5pt},
    xmin = -8,
    xmax = 8,
    domain=-8:8
]
\addplot[blue,mark=none,samples=500]   (x,{sigma(x/1.0)});\addlegendentry{$g(u)$}
\addplot[red,mark=none,samples=500]   (x,{sigmap(x/1.0)});\addlegendentry{$\frac{dg(u)}{du}$}
\end{axis}
\end{tikzpicture}
\hspace{-5pt}
{
\begin{tikzpicture}[declare function={sigma(\x)=1/(1+exp(-\x));sigmap(\x)=sigma(\x)*(1-sigma(\x));}]
\begin{axis}[%
	width=0.57\linewidth,
	height=0.5\linewidth,
	xlabel={$u= s_{qz}-s_{qx}$ },
	title={$g(u)= \sigma_{\tau_2}(u), \tau_2 =0.01$},
	legend cell align={left},
	legend pos=north west,
    legend style={cells={anchor=east}, font =\small, fill opacity=0.8, row sep=-2.5pt},
    xmin = -1,
    xmax = 1,
    domain=-1:1
]
\addplot[blue,mark=none,samples=500]   (x,{sigma(x/0.01)});
\addplot[red,mark=none,samples=500]   (x,{sigmap(x/0.01)});
\end{axis}
\end{tikzpicture}
}
\caption{The two sigmoid functions which replace the Heaviside step function for counting the positive examples in the short-list of size $k$ (left) and for estimating the rank of examples (right).
\label{fig:sigmoids}
}
\end{center}
\end{figure}

\begin{figure}[t]
\begin{center}
\begin{tikzpicture}[declare function={sigma(\x)=1/(1+exp(-\x));sigmap(\x)=sigma(\x)*(1-sigma(\x));}]
\begin{axis}[%
	width=0.99\linewidth,
	height=0.5\linewidth,
	xlabel={$r$},
	ylabel={$\frac{d\sigma_{\tau_1}(u)}{du},~u=k-r$},
	legend cell align={left},
	legend pos=north east,
    legend style={cells={anchor=east}, font =\tiny, fill opacity=0.8, row sep=-2.5pt},
    xtick = {1,3,5,7,9},
    xmin = 1,
    xmax = 10,
    domain=0:30
]
\addplot[blue,mark=none,samples=500]   (x,{sigmap(x-1)});\addlegendentry{$k=1$}
\addplot[red,mark=none,samples=500]   (x,{sigmap(x-2)});\addlegendentry{$k=2$}
\addplot[green,mark=none,samples=500]   (x,{sigmap(x-4)});\addlegendentry{$k=4$}
\addplot[black,mark=none,samples=500]   (x,{sigmap(x-1)+sigmap(x-2)+sigmap(x-4)});\addlegendentry{$K=\{1,2,4\}$};
\end{axis}
\end{tikzpicture}
\caption{Gradient magnitude of the sigmoid used to count the positive examples in the short-list of size $k$ versus the rank $r$ (equal to $r_\Omega(q,x)$, see \equ{rank}) of a positive example $x$. It shows how much a positive example is pushed towards lower ranks depending on its current rank. In the case of multiple values for $k$, the total gradient is equivalent to the sum of the separate ones.
\label{fig:deriv}}
\end{center}
\vspace{-2em}
\end{figure}
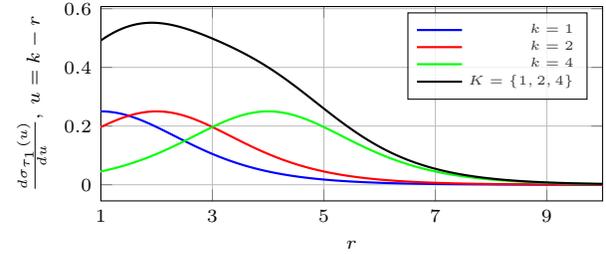

\paragraph{Recall@k surrogate loss.} The computation of recall in \equ{recall_full} involves the use of the Heaviside step function. The gradient of the Heaviside step function is a Dirac delta function. Hence, direct optimization of recall with back-propagation is not feasible. A common smooth approximation of the Heaviside step function is provided by the logistic function~\cite{km+15,ikm+15,ikm+17}, a common sigmoid function $\sigma_{\tau}: \real \rightarrow \real$ controlled by temperature $\tau$, which is given by
\begin{equation}
\sigma_\tau(u) = \frac{1}{1+e^{-\frac{u}{\tau}}},
\end{equation}
where large (small) temperature value leads to worse (better) approximation and denser (sparser) gradient. This approximation is common in the machine learning literature for several tasks~\cite{sh+09,gls+16,mmt+17} and also appears in the approximation of the Average Precision evaluation metric~\cite{bxk+20}, which is used for the same task as ours. By replacing the step function with the sigmoid function, a smooth approximation of recall is obtained as
\begin{equation}
     \tilde{R}_\Omega^k(q) = \frac{\sum\limits_{x \in P_q} \sigma_{\tau_1} (k  - 1 - \sum\limits_{\substack{z \in \Omega\\ z \neq x}} \sigma_{\tau_2}(s_{qz} - s_{qx}) )}{|P_q|},
\label{equ:smooth_recall}
\end{equation}
which is differentiable and can be used for training with back-propagation. The two sigmoids have different function domains and, therefore, different temperatures (see Figure~\ref{fig:sigmoids}). The minimized single-query loss in a mini-batch $B$, with size $M=|B|$, and query $q\in B$ is given by 
\begin{equation}
L^k(q) = 1- \tilde{R}_{B\setminus q}^k(q).
\end{equation}
while incorporation of multiple values of $k$ is performed in the loss given by 
\begin{equation}
L^K(q) = \frac{1}{|K|}\sum_{k \in K} L^k(q).
\label{equ:loss}
\end{equation}
Figure~\ref{fig:deriv} shows the impact of using single or multiple values for $k$. 

All examples in the mini-batch are used as queries and the average loss over all queries is minimized during the training.
The proposed loss is referred to as \emph{Recall@k Surrogate loss}, or RS@k loss for brevity.

To allow for 0 loss when $k$ is smaller than the number of positives (note that exact recall@k is less than 1 by definition), we slightly modify \equ{smooth_recall} during the training. Instead of dividing by $|P_q|$, we divide by $\min(k, |P_q|)$, and, consequently, we clip values larger than $k$ in the numerator to avoid negative loss values.

\paragraph{Similarity mixup (SiMix).} Given original batch $B$, virtual batch $\hat{B}$ is created by mixing all pairs of positive examples in the original batch.
Embeddings of examples $x\in B$ and $z \in B$ are used to generate mixed embedding
\begin{equation}
    \vv_{xz\alpha} = \alpha \vx + (1-\alpha) \vz \quad | \quad \alpha \sim U(0,1),
\end{equation}
for a virtual example that is denoted by $xz\alpha \in \hat{B}$.
The similarity of an original example $w\in B$ to the virtual example $xz\alpha \in \hat{B}$ is given by
\begin{equation}
s(w,xz\alpha) = \vw^\top \vv_{xz\alpha} = \alpha s_{wx} + (1-\alpha) s_{wz}, 
\label{equ:simix1} 
\end{equation}
where the original and virtual examples can be the query and database examples, respectively, or vice versa. 
In case both examples are virtual, \eg $xz\alpha_1 \in \hat{B}$ used as a query and $yw\alpha_2 \in \hat{B}$ as a part of the database, then their similarity is given by
\begin{align}
s(xz\alpha_1,yw\alpha_2) &= \vv_{xz\alpha_1}^\top \vv_{yw\alpha2} \nonumber\\ 
        &= \alpha_1 \alpha_2 s_{xy} + (1-\alpha_1) (1- \alpha_2) s_{zw} \nonumber\\
        &+ \alpha_1 (1-\alpha_2) s_{xw}+(1-\alpha_1) \alpha_2 s_{zy}.
\label{equ:simix2} 
\end{align}
The pairwise similarities that appear on the right-hand side of the previous formulas, \eg $s_{wx}$ and $s_{wz}$ in \equ{simix1}, are computed from the embeddings of the original, non-virtual examples and are also required for the computation of the RS@k without any virtual examples. Therefore, the mini-batch is expanded to $B \cup \hat{B}$ by adding virtual examples without the need for explicit construction of the corresponding embeddings or computation of the similarity via dot product; simple mixing of the corresponding pairwise scalar similarities is enough. SiMix reduces to mixing pairwise similarities due to the lack of re-normalization of the mixed embeddings, which is different to existing practice in prior work~\cite{vpa+21,gk+20,gkk+21,ksp+20} and brings training efficiency benefits.

Virtual examples are created only between examples of the same classes and are labeled according to the class of the original examples that are mixed. Virtual examples are used both as queries and as database examples, while mixing is applied to all pairs of positive examples inside a mini-batch.

\paragraph{Overview.} An overview of the training process with the proposed loss and SiMix is given in Algorithm \ref{alg:main}. In case SiMix is not used, then lines \ref{lin:vbatch}, \ref{lin:vsim1}, \ref{lin:vsim2} and \ref{lin:expand} are skipped. It is assumed that each image in training is labeled to a class. Mini-batches of size $M$ are generated by randomly sampling $m$ images per class out of $\nicefrac{M}{m}$ sampled classes.

\begin{algorithm}
\caption{Training with RS@k and SiMix.\label{alg:main}}
\footnotesize
\algrenewcommand\algorithmicindent{0.5em}%
\begin{algorithmic}[1]
\Procedure{Train-RS@k}{$X$, $Y$, $M$, $m$}
\color{gray}
\State $X: $ training images
\State $Y: $ class labels
\State $M: $ mini-batch size
\State $m: $ number of images per class in mini-batch
\color{black}
\State
\State $\theta \gets$ initialize according to pre-training \comment{use ImageNet}
\For{$\text{iteration} \in [1,\ldots, \text{number-of-iterations}]$}
\State $loss \gets 0$ \comment{set batch loss to zero}
\State $B \gets$ \Call{Batch-sampler}{$X$, $Y$, $M$, $m$} 
\State $\hat{B} \gets$ \Call{Virtual-batch}{$B$} \comment{enumerate virtual examples} \label{lin:vbatch}
\State \algorithmicfor { $(x,z) \in B \times B$} \algorithmicdo { compute $s(x,z)$ }  \comment{use $\vx^\top \vz$} 
\State \algorithmicfor { $(x,z) \in B \times \hat{B}$} \algorithmicdo { compute $s(x,z)$ }  \comment{use \equ{simix1}} \label{lin:vsim1}
\State \algorithmicfor { $(x,z) \in \hat{B} \times \hat{B}$} \algorithmicdo { compute $s(x,z)$ }  \comment{use \equ{simix2}} \label{lin:vsim2}
\State $B \gets B \cup \hat{B}$ \comment {expand batch with virtual examples} \label{lin:expand}
\For{$q \in B$} \comment {use each image in the batch as query}
\State $loss \gets loss + L^K(q)$ \comment {Recall@k loss \equ{loss}}
\EndFor
\State $\theta \gets $ \Call{Minimize}{$\frac{loss}{|B|}$} \comment {SGD update}
\EndFor
\EndProcedure
\end{algorithmic}
\end{algorithm}

\section{Experiments}
\label{sec:experiments}

\subsection{Datasets}
\label{sec:datasets}

\begin{table}
\setlength\extrarowheight{1pt}
\begin{center}
\small
\begin{tabular}{l|r|r|r}
    \hline
    \textbf{Dataset} & \textbf{\#Images} &  \textbf{\#Classes} & \textbf{\#Avg}\\
    \hline\hline
    iNaturalist Train~\cite{vms+18} & $325,846$ & $5,690$ & $57.3$ \\
    iNaturalist Test~\cite{vms+18} & $136,093$ & $2,452 $ & $55.5$ \\
    VehicleID Train~\cite{ltw+16} & $110,178$ & $13,134$ & $8.4$ \\
    VehicleID Test~\cite{ltw+16} & $40,365 $ & $4,800$ & $8.4$ \\
    SOP Train~\cite{ohb16} & $59,551 $ &  $11,318$ & $5.3$ \\
    SOP Test~\cite{ohb16} & $60,502$ &  $11,316$ & $5.3$ \\
    Cars196 Train~\cite{ksd+13} & $8,054$ & $98$ & $82.1$ \\
    Cars196 Test~\cite{ksd+13} & $8,131$ & $98$ & $82.9$ \\
    \hline
    $\mathcal{R}$Oxford~\cite{rit+18} & $4,993$ & $11$ & n/a \\
    $\mathcal{R}$Paris~\cite{rit+18} & $6,322$ & $11$ & n/a \\
    GLDv1~\cite{nas+17} & $1,060,709$ & $12,894$ & $82.3$ \\
    \hline
\end{tabular}
\end{center}
\vspace{-1.5em}
\caption{Dataset composition for training and evaluation.}
\label{tab:datasets}
\vspace{-1em}
\end{table}

The training and evaluation is performed on four widely used image retrieval benchmarks, namely iNaturalist~\cite{vms+18}, PKU VehicleID~\cite{ltw+16}, Stanford Online Products~\cite{ohb16} (SOP) and Stanford Cars~\cite{ksd+13} (Cars196). Recall at top $k$ retrieved images,  denoted by r@k, is one of the standard evaluation metrics in these benchmarks.  Metric r@k is 1 if at least one positive image appears in the top $k$ list, otherwise 0. The metric is averaged across all queries. Note that this is different from the standard definition of recall in \equ{recall}. 

iNaturalist~\cite{vms+18} is firstly used by Brown~\etal~\cite{bxk+20}, whose setup we follow: $5,690$ classes for training and $2,452$ classes for testing. For VehicleID, according to the standard setup~\cite{ltw+16}, $13,134$ classes are used for training, and the evaluation is conducted on the predefined small ($800$ classes), medium ($1600$ classes) and large ($2400$ classes) test sets. For SOP~\cite{ohb16} and Cars196~\cite{ksd+13}, the standard experimental setup of Song~\etal~\cite{sxj+15} is followed. The first half of the classes are used for training and the rest for testing, resulting in $11,318$ classes for SOP and $98$ for Cars196.

The method is evaluated for instance-level search on Revisited Oxford ($\mathcal{R}$Oxford) and Paris ($\mathcal{R}$Paris) benchmark~\cite{rit+18}, where the evaluation metric is mean Average Precision (mAP). The training uses the Google Landmarks dataset (GLDv1)~\cite{nas+17} to perform a comparison with the work of Revaud~\etal~\cite{rar+19} and their AP loss. The validation is performed according to the work of Tolias~\etal~\cite{tjc20}. 

The number of examples, classes, and average number of examples per class can be found in Table. \ref{tab:datasets}. Note that these datasets are diverse in the number of training examples, the number of classes, and the number of examples per class, ranging from class balanced~\cite{ksd+13} to long-tailed~\cite{vms+18}.

\subsection{Implementation details}
\label{sec:implementation_details}

Implementation details are identical for the four image retrieval benchmarks but differ for  $\mathcal{R}$Oxford/$\mathcal{R}$Paris to follow and compare to prior work~\cite{rar+19}. Differences are clarified when needed.

\paragraph{Architecture.} An ImageNet~\cite{dsl+09} pre-trained ResNet-50~\cite{hzr+16} is used as the backbone for deep image embeddings. Building on the standard implementation of~\cite{rms+20}, the BatchNorm parameters are kept frozen during the training. After the convolutional layers, Generalized mean pooling~\cite{rtc19} and layer normalization~\cite{bkh+16} are used,  similar to~\cite{tdt20}. For vision transformers~\cite{dbk+21} ViT-B/32 and ViT-B/16 with an ImageNet-21k initialization from the timm library~\cite{rw2019timm} are used. The last layer of the model is a $d$ dimensional fully connected (FC) layer with $L_{2}$ normalization. In the case of $\mathcal{R}$Oxford/$\mathcal{R}$Paris, ResNet-101~\cite{hzr+16} is used, layer normalization is not added, while the FC layer is initialized with the result of whitening~\cite{rtc19}.

\paragraph{Training hyper-parameters.} For ResNet architectures, Adam optimizer~\cite{kb15} is used and for vision transformers, AdamW~\cite{lh+19} is used. This paper follows the standard class-balanced-sampling~\cite{mbl20,bxk+20,tdt20} with $4$ samples per class for all the datasets, while classes with less than $4$ samples are not used for training. Unless stated otherwise, the batch size for training is set $4,000$ for all datasets but Cars196 where it is equal to $4\times\#\text{classes}=392$. Following the setup of ProxyNCA++~\cite{tdt20}, the training set is split into training and validation by using the first half of the classes for training and the other half for validation. With this split, a grid search determines the learning rate, decay steps, decay size and the total number of epochs. Once the hyper-parameters are fixed, training is conducted once on the entire training set and evaluated on the test set. When training on GLDv1 and testing on $\mathcal{R}$Oxford/$\mathcal{R}$Paris, the batch size is set to $4096$~\cite{rar+19}, and training is performed for 500 batches, while other training hyper-parameters are set as in the work and GitHub implementation of Radenovic \etal~\cite{rtc19}. Note that the hyper-parameters for each dataset will be released with the implementation.

\paragraph{RS@k hyper-parameters.} The proposed Recall@k Surrogate (RS@k) loss \equ{smooth_recall} contains three hyper-parameters: sigmoid temperature $\tau_{2}$ - applied on similarity differences, sigmoid temperature $\tau_{1}$ - applied on ranks and the set of values for $k$ for which the loss is computed. Both sigmoid temperatures are kept fixed across all the experiments as $\tau_{2}=0.01$ (same as~\cite{bxk+20}) and $\tau_{1}=1$. The values of $k$ are kept fixed as $k=\{1,2,4,8,16\}$ without SiMix and $k=\{1,2,4,8,12,16,20,24,28,32\}$ with SiMix. For GLDv1~\cite{nas+17}, this is $k=\{1,2,4\}$, and $k=\{1,2,4,8\}$, respectively. The values of $k$ are studied in the supplementary materials and the sigmoid temperature $\tau_{1}$ are investigated in Section \ref{sec:effect_hyperparams}, where it is observed that the method is not very sensitive to these hyper-parameters. 

\paragraph{Large batch size.} To dispense with the GPU hardware constraints and manage to train with the large batch size, we follow the multistage back-propagation of Revaud~\etal~\cite{rar+19}. A forward pass is performed to obtain all embeddings while intermediate tensors are discarded from memory. Then, the loss is computed, and so are the gradients \wrt the embeddings. Finally, each of the embeddings is recomputed, this time allowing the propagation of the gradients. Note that there is no implementation online of this approach and that the code of this work will become publicly available. Algorithm \ref{alg:main} does not include such implementation details, but it is compatible with such an extension. The batch-size impact for the proposed RS@k loss function is validated in Section \ref{sec:effect_hyperparams}.

\paragraph{Discussion.} The methods in the literature use different embedding sizes, $d$, therefore, the models for the RS@k loss are trained with two embedding sizes of $d=128$ and $d=512$ for image retrieval benchmarks \cite{vms+18,ltw+16,ohb16,ksd+13}, and $d=2048$ for $\mathcal{R}$Oxford/$\mathcal{R}$Paris \cite{rit+18}, to allow a fair comparison. In the standard split, the image retrieval benchmarks~\cite{ksd+13,ohb16,ltw+16,vms+18} do not contain an explicit validation set; as a result, image retrieval methods often tune the hyper-parameters on the test set, leading to the issue of training with test set feedback. This issue has been studied in~\cite{mbl20}, which proposes to train different methods with identical hyper-parameters. The setup of~\cite{mbl20} is not directly usable for experiments with the RS@k loss, as large batch sizes are crucial to estimate recall@k accurately. Furthermore, their setup does not allow mixup. Therefore, instead of following~\cite{mbl20}, the issue is eliminated by using a part of the training set for validation as described above.

\subsection{Evaluation}
\label{sec:evaluation}

\begin{table*}[t]
\begin{center}
\footnotesize
\setlength{\tabcolsep}{1.5pt}
    \setlength\extrarowheight{-2pt}
    \begin{tabular}{@{\zsp}l@{\xssp}l@{\xssp}|cccc|cccc|cc|cc|cc|cccc}
    \hline
    \multirow{3}[1]{*}{Method}& \multirow{3}[1]{*}{Arch.$^\text{dim}$} & \multicolumn{4}{c|}{iNaturalist \cite{vms+18}} & \multicolumn{4}{c|}{SOP \cite{ohb16}} & \multicolumn{6}{c|}{VehicleID \cite{ltw+16}} & \multicolumn{4}{c}{Cars196 \cite{ksd+13}}\\
    \cline{3-20}
    & & \multicolumn{15}{c}{r@k}\\ \cline{3-20}
    & & & & & & & & & & \multicolumn{2}{c|}{Small} & \multicolumn{2}{c|}{Medium} & \multicolumn{2}{c|}{Large} & & & & \\
    & & 1 & 4 & 16 & 32 & $10^0$ & $10^1$ & $10^2$ & $10^3$ & 1 & 5 & 1 & 5 & 1 & 5 & 1 & 2 & 4 & 8\\
    \hline \hline
    
	ProxyNCA \cite{mtl+17} & {\scriptsize$I_{1}^{128}$} &
	$\underline{61.6}$ & 
	$\underline{77.4}$ & 
	$\underline{87.0}$ & 
	$90.6$ &

	$73.7$ & 
	- & 
	- & 
	- &
	
	- &
	- &
	- &
	- &
	- &
	- &
	
	$73.2$ &
	$82.4$ &
	$86.4$ &
	$88.7$
	\\
	
	Margin \cite{wms+17} & {\scriptsize$R_{50}^{128}$} &
	$58.1$ & 
	$75.5$ & 
	$86.8$ & 
	$\underline{90.7}$ &

	$72.7$ &
	$86.2$ & 
	$93.8$ & 
	$98.0$ &
	
	- &
	- &
	- &
	- &
	- &
	- &
	
	$\underline{79.6}$ &
	$\underline{86.5}$ &
	$\underline{91.9}$ &
	$\underline{95.1}$
	\\
	
	Divide \cite{skp+15} & {\scriptsize$R_{50}^{128}$} &
	- &
	- &
	- &
	- &
	
	$75.9$ & 
	$88.4$ & 
	$94.9$ &
	$98.1$ &

    $87.7$ &
    $92.9$ &
    $85.7$ &
    $90.4$ &
    $82.9$ &
    $90.2$ &
    
    - &
    - &
    - &
    -
	\\
	
	MIC \cite{rbo+19} & {\scriptsize$R_{50}^{128}$} &
	- &
	- &
	- &
	- &

	$77.2$ &
	$89.4$ &
	$95.6$ &
	-  &

    $86.9$ &
    $93.4$ &
    - &
    - &
    $82.0$ &
    $91.0$ &
    
    - &
    - &
    - &
    -

	\\
	
	Cont. w/M \cite{wzh+20} & {\scriptsize$R_{50}^{128}$} &
	- &
	- &
	- &
	- &
	
	$\underline{80.6}$ &
	$\underline{91.6}$ &
	$\underline{96.2}$ &
	$\underline{98.7}$ &

    $\underline{94.7}$ &
    $\underline{96.8}$ &
    $\underline{93.7}$ &
    $\underline{95.8}$ &
    $\underline{93.0}$ &
    $\underline{95.8}$ &
    
    - &
    - &
    - &
    -
	\\
	
	\hline
	RS@k\textsuperscript{\dag} & {\scriptsize$R_{50}^{128}$} &
	$69.3$  &
	$82.9$  &
	$90.6$  &
	$93.1$ &
	
	$80.6$ &
	$91.6$ &
	$96.4$ &
	$\bm{98.8}$ &

    $\bm{95.6}$ &
    $\bm{97.8}$ &
    $\bm{94.4}$ &
    $\bm{96.8}$ &
    $\bm{93.5}$ &
    $\bm{96.6}$ &
    
    $78.1$ &
    $85.8$ &
    $91.1$ &
    $94.5$
	\\

%
%
%
	
	RS@k\textsuperscript{\dag} +SiMix & {\scriptsize$R_{50}^{128}$} &
	$\bm{69.6}$ & 
	$\bm{83.3}$ & 
	$\bm{91.2}$ & 
	$\bm{93.8}$ &
	
	$\bm{80.9}$ &
	$\bm{91.7}$ &
	$\bm{96.5}$ &
	$\bm{98.8}$ &

    $95.4$ &
    $97.5$ &
    $93.8$ &
    $96.6$ &
    $93.0$ &
    $96.2$ &
    
    $\bm{84.7}$ &
    $\bm{90.9}$ &
    $\bm{94.7}$ &
    $\bm{96.9}$ 
	\\

	& & 
	\tiny{\textcolor{blue}{$+21\%$}} &
	\tiny{\textcolor{blue}{$+26\%$}} &
	\tiny{\textcolor{blue}{$+32\%$}} &
	\tiny{\textcolor{blue}{$+33\%$}} &
	
	\tiny{\textcolor{blue}{$+1.5\%$}} &
	\tiny{\textcolor{blue}{$+1.2\%$}} &
	\tiny{\textcolor{blue}{$+7.9\%$}} &
	\tiny{\textcolor{blue}{$+7.7\%$}} &

	\tiny{\textcolor{blue}{$+17\%$}} &
	\tiny{\textcolor{blue}{$+31\%$}} &
	\tiny{\textcolor{blue}{$+11\%$}} &
	\tiny{\textcolor{blue}{$+24\%$}} &
	\tiny{\textcolor{blue}{$+7.1\%$}} &
	\tiny{\textcolor{blue}{$+19\%$}} &

	\tiny{\textcolor{blue}{$+25\%$}} &
	\tiny{\textcolor{blue}{$+33\%$}} &
	\tiny{\textcolor{blue}{$+35\%$}} &
	\tiny{\textcolor{blue}{$+37\%$}}
	\\

%
%
%

    \hline \hline
	FastAP \cite{chx+19} & {\scriptsize$R_{50}^{512}$} &
	$60.6$ & 
	$77.0$ & 
	$87.2$ & 
	$90.6$ &

	$76.4$ &
	$89.0$ & 
	$95.1$ & 
	$98.2$ &

    $91.9$ &
    $96.8$ &
    $90.6$ &
    $95.9$ &
    $87.5$ &
    $95.1$ &
    
    - &
    - &
    - &
    -
	\\
	
	MS \cite{whh+19} & {\scriptsize$I_{3}^{512}$} &
	- &
	- &
	- &
	- &
	
	$78.2$ &
	$90.5$ &
	$96.0$ &
	$98.7$ &

	- &
	- &
	- &
	- &
	- &
	- &
	
	$84.1$ &
	$90.4$ &
	$94.0$ &
	$96.1$
	\\
	
	NormSoftMax \cite{zw18} & {\scriptsize$R_{50}^{512}$} &
	- &
	- &
	- &
	- &
	
	$78.2$ &
	$90.6$ &
	$96.2$ &
	- &

	- &
	- &
	- &
	- &
	- &
	- &
	
	$84.2$ &
	$90.4$ &
	$94.4$ &
	$96.9$
	\\
	
	Blackbox AP \cite{rmp+20} & {\scriptsize$R_{50}^{512}$} &
	$62.9$ &
	$79.0$ &
	$88.9$ &
	$92.1$ &
	
	$78.6$ &
	$90.5$ &
	$96.0$ &
	$98.7$ &
	
	- &
	- &
	- &
	- &
	- &
	- &

    - &
    - &
    - &
    -
	\\
	
	Cont. w/M \cite{wzh+20} & {\scriptsize$I_{3}^{512}$} &
	- &
	- &
	- &
	- &
	
	$79.5$ &
	$90.8$ &
	$96.1$ &
	$98.7$ &

    $94.6$ &
    $96.9$ &
    $\underline{93.4}$ &
    $96.0$ &
    $\underline{93.0}$ &
    $96.1$ &

    - &
    - &
    - &
    -
	\\
	
	HORDE \cite{jph+19} & {\scriptsize$R_{50}^{512}$} &
	- &
	- &
	- &
	- &
	
	$80.1$ &
	$91.3$ &
	$96.2$ &
	- &

	- &
	- &
	- &
	- &
	- &
	- &
	
	$86.2$ &
	$91.9$ &
	$95.1$ &
	$97.2$
	\\
	
	ProxyNCA++ \cite{tdt20} & {\scriptsize$R_{50}^{512}$} &
	- &
	- &
	- &
	- &
	
	$\underline{80.7}$ &
	$\underline{92.5}$ &
	$96.7$ &
	$98.9$ &

	- &
	- &
	- &
	- &
	- &
	- &
	
	$\underline{86.5}$ &
	$\underline{92.5}$ &
	$\underline{95.7}$ &
	$\underline{\bm{97.7}}$
	\\

	SAP \cite{bxk+20} & {\scriptsize$R_{50}^{512}$} &
	$\underline{67.2}$ &
	$\underline{81.8}$ &
	$\underline{90.3}$ &
	$\underline{93.1}$ &
	
	$80.1$ &
	$91.5$ &
	$\underline{96.6}$ & 
	$\underline{99.0}$ &

    $\underline{94.9}$ &
    $\underline{97.6}$ &
    $93.3$ &
    $\underline{96.4}$ &
    $91.9$ &
    $\underline{96.2}$ &
    
    $76.1$ &
    $84.3$ &
    $89.8$ &
    $93.8$
	\\
	
	SAP\textsuperscript{\dag} \cite{bxk+20} {\tiny +GeM +LN} & {\scriptsize$R_{50}^{512}$} &
	$68.7$ &
	$82.7$ &
	$90.9$ &
	$93.5$ &
	
	$80.3$ &
	$92.0$ &
	$96.9$ &
	$99.0$ &

    $94.2$ &
    $97.2$ &
    $92.7$ &
    $96.2$ &
    $91.0$ &
    $95.8$ &
    
    $78.2$ &
    $85.6$ &
    $90.8$ &
    $94.3$
	\\

	

    

	
	
	

	
    
	

	\hline
	
	

	RS@k\textsuperscript{\dag} & {\scriptsize$R_{50}^{512}$} &
	$71.2$ &
	$84.0$ &
	$91.3$ &
	$93.6$ &
	
	$\bm{82.8}$ & 
	$\bm{92.9}$ & 
	$\bm{97.0}$ & 
	$99.0$ &
	
    $\bm{95.7}$ &
    $\bm{97.9}$ &
    $\bm{94.6}$ &
    $\bm{96.9}$ &
    $\bm{93.8}$ &
    $\bm{96.6}$ &
    
    $80.7$ &
    $88.3$ &
    $92.8$ &
    $95.7$
	\\

%
%
%

	RS@k\textsuperscript{\dag} +SiMix & {\scriptsize$R_{50}^{512}$} &
	$\bm{71.8}$ &
	$\bm{84.7}$ &
	$\bm{91.9}$ &
	$\bm{94.3}$ &
	
	$82.1$ & 
	$92.8$ &
	$\bm{97.0}$ &
	$\bm{99.1}$ &

    $95.3$ &
    $97.7$ &
    $94.2$ &
    $96.5$ &
    $93.3$ &
    $96.4$ &
    
    $\bm{88.2}$ &
    $\bm{93.0}$ &
    $\bm{95.9}$ &
    $97.4$
	\\
	
	& & 
	\tiny{\textcolor{blue}{$+14\%$}} &
	\tiny{\textcolor{blue}{$+16\%$}} &
	\tiny{\textcolor{blue}{$+16\%$}} &
	\tiny{\textcolor{blue}{$+17\%$}} &
	
	\tiny{\textcolor{blue}{$+11\%$}} &
	\tiny{\textcolor{blue}{$+5.3\%$}} &
	\tiny{\textcolor{blue}{$+12\%$}} &
	\tiny{\textcolor{blue}{$+10\%$}} &

	\tiny{\textcolor{blue}{$+16\%$}} &
	\tiny{\textcolor{blue}{$+13\%$}} &
	\tiny{\textcolor{blue}{$+18\%$}} &
	\tiny{\textcolor{blue}{$+14\%$}} &
	\tiny{\textcolor{blue}{$+11\%$}} &
	\tiny{\textcolor{blue}{$+10\%$}} &

	\tiny{\textcolor{blue}{$+13\%$}} &
	\tiny{\textcolor{blue}{$+6.7\%$}} &
	\tiny{\textcolor{blue}{$+4.7\%$}} &
	\tiny{\textcolor{red}{$-13\%$}}
	\\

%
%
%
%

    \hline \hline
    
    SAP\textsuperscript{\dag} \cite{bxk+20} & {\scriptsize ViT-B/32$^{512}$} &
    $72.2$ &
	$84.6$ &
	$91.6$ &
	$93.9$ &
	
    $83.7$ &
	$94.0$ &
	$97.8$ &
	$99.3$ &
	
	$94.8$ &
	$97.7$ &
	$93.5$ &
	$96.8$ &
	$92.1$ &
	$96.3$ &
	
	$78.1$ &
	$85.7$ &
	$91.0$ &
	$94.8$
	\\

    RS@k\textsuperscript{\dag} & {\scriptsize ViT-B/32$^{512}$} & 
    $75.9$ & 
    $87.1$ &
    $93.1$ &
    $95.1$ &

    $85.1$ & 
    $94.6$ &
    $98.0$ &
    $99.3$ &

    $95.1$ & 
    $97.7$ &
    $94.1$ &
    $96.7$ &
    $93.2$ &
    $96.5$ &
    
    $78.1$ &
    $86.4$ &
    $92.3$ &
    $95.6$
    \\

%
%
    
	\hline
	
    SAP\textsuperscript{\dag} \cite{bxk+20} & {\scriptsize ViT-B/16$^{512}$} &
	$79.1$ &
	$89.0$ &
	$94.2$ &
	$95.8$ &
	
	$86.6$ &
	$95.4$ &
	$98.4$ &
	$99.5$ &

	$95.5$ &
	$97.7$ &
	$94.2$ &
	$96.9$ &
	$93.1$ &
	$96.6$ &
	
	$86.2$ &
	$92.1$ &
	$95.1$ &
	$97.2$
	\\

    RS@k\textsuperscript{\dag} & {\scriptsize ViT-B/16$^{512}$} & 
    $83.9$ & 
    $92.1$ &
    $95.9$ &
    $97.2$ &
    
    $88.0$ & 
    $96.1$ &
    $98.6$ &
    $99.6$ &
    
    $96.2$ & 
    $98.0$ &
    $95.2$ &
    $97.2$ &
    $94.7$ &
    $97.1$ &
    
    $89.5$ &
    $94.2$ &
    $96.6$ &
    $98.3$
    \\

%
%
    \hline
    \end{tabular}
    \vspace{-1em}
    \caption{Recall@$k(\%)$ on iNaturalist~\cite{vms+18}, Stanford Online Products (SOP)~\cite{ohb16}, PKU VehicleID~\cite{ltw+16} and Stanford Cars (Cars196)~\cite{ksd+13}. Best results are shown with \textbf{bold}, previous state-of-the-art with \underline{underline} and relative gains over the state-of-the-art in \% of error reduction with \textcolor{blue}{blue} and relative declines in \textcolor{red}{red}. Methods marked with $\dag$ were trained using the same pipeline by the authors of this paper.}
    \label{tab:metriclearning}
    \vspace{-2em}
\end{center}
\end{table*}

Unless otherwise stated, the results of the competing methods are taken from the original papers. Methods marked with a $\dag$ were trained by the authors of this paper, using the same implementation as used for the RS@k loss. The results on image retrieval benchmarks~\cite{ksd+13,ohb16,ltw+16,vms+18} are compared with the methods that use either ResNet-50~\cite{hzr+16} or Inception network~\cite{slj+15}. ResNet-50~\cite{hzr+16} is represented as $R_{50}^{d}$ in the tables and the standard Inception network~\cite{slj+15} as $I_{1}^{d}$, the Inception network with BatchNorm as  $I_{3}^{d}$ (same as ~\cite{tdt20}). Here $d$ is the embedding size. On all the datasets, the performance of the baseline, Smooth-AP (SAP)~\cite{bxk+20}, is also reported with Generalized mean pooling~\cite{rtc19} and layer normalization~\cite{bkh+16}, shown as SAP\textsuperscript{\dag} (+Gem +LN). This is to eliminate any performance boost in the comparisons that were caused by the architecture. Note that unless otherwise stated in our experiments, the batch size for SAP is set as $384$, the same as the original implementation~\cite{bxk+20}. Further, we demonstrate the performance of SAP and RS@k on ViT-B architectures. The variant of ViT-B that uses a patch size of $32\times32$ is denoted by ViT-B/32 and the one that uses a patch size of $16\times16$ by ViT-B/16.

\paragraph{iNaturalist.} The results on iNaturalist~\cite{vms+18} species recognition are presented in Table \ref{tab:metriclearning}. The performances of the competing methods are taken from~\cite{bxk+20}, which uses the official implementations of these methods. It can be clearly seen that the RS@k outperforms classification and pairwise losses, including the three AP approximation losses, reaching the recall@1 score of $71.8\%$ with SiMix, an error reduction of $14\%$.

\paragraph{SOP.} The performance on SOP~\cite{ohb16} is presented in Table \ref{tab:metriclearning}, along with the comparisons with the competing methods. The proposed RS@k loss demonstrates clear state-of-the-art results, surpassing ProxyNCA++~\cite{tdt20} by $2.0\%$ on recall@1, an error reduction of $10.4\%$. If a smaller batch size, equal to $384$, is used for RS@k, it reaches a performance of $81.2\%$, $92.2\%$, $96.9\%$ and $99.0\%$ on r@$10^{0}$, r@$10^{1}$, r@$10^{2}$ and r@$10^{3}$ respectively. This result shows that large batch size helps in improving the performance, but RS@k outperforms the competing methods even with smaller batch size.

\paragraph{VehicleID.} The results on VehicleID~\cite{ltw+16} are presented in Table \ref{tab:metriclearning}. RS@k outperforms the competing methods both with and without SiMix. Better results were observed without SiMix where RS@k reaches recall@1 performance of $95.7\%$, $94.6\%$ and $93.8\%$ on the small, medium, and large test sets, respectively.

\begin{table*}[t!]
\tablestyle{3pt}{1.3}
\begin{center}
\setlength\extrarowheight{-2pt}
\newcommand{\xdagger}{^{\dagger}}
\def\arraystretch{1.0}
\small
\begin{tabular}	{l@{\msp}r@{\msp}r@{\msp}r@{\msp}c@{\msp}c@{\msp}c@{\msp}c@{\msp}c@{\msp}c@{\msp}c@{\msp}c@{\msp}c@{\msp}c}
\hline
\multirow{2}{*}{Arch.}        &    \multirow{2}{*}{Loss} & \multirow{2}{*}{Train-set} & \multirow{2}{*}{} & \multicolumn{2}{c}{Mean}  &   \multicolumn{2}{c}{\ro} & \multicolumn{2}{c}{\ro\hspace{-3pt}+\rdis} & \multicolumn{2}{c}{\rpa} & \multicolumn{2}{c}{\rp\hspace{-3pt}+\rdis}     \\ \cline{5-14}
															 & &&    												     & all & \rdis &   med  &   hard         &      med  &   hard          &    med  &   hard      &           med  &   hard         \\ \hline \hline
GeM$\ast$                                         & AP \cite{hls18} & Landmarks-clean~\cite{bsc+14}\cite{gar+17} & ~\cite{rar+19}/~\cite{tjc20}    & 49.7 & 36.7 & 67.1 & 42.3 & 47.8 & 22.5 & 80.3 & 60.9 & 51.9 & 24.6 \\             
GeM$\ast$                                         & AP \cite{hls18} & GLDv1~\cite{nas+17} & ~\cite{rar+19}/github    & - & -  & 66.3 & 42.5 & - & -  & 80.2 & 60.8  & -  & - \\             
GeM\dag                & SAP \cite{bxk+20} & GLDv1~\cite{nas+17} & ~\cite{bxk+20}    & 52.7 & 40.6 & 67.9 & 46.3 &  49.5 & 25.8 & 81.7 & 63.3 & 57.4 & 29.8 \\     
GeM\dag                & RS@k & GLDv1~\cite{nas+17} & ours    & 53.1 & 41.0 & 68.3 & 46.1 & 50.1 & 25.8 & 82.1 & 63.9 & 57.9 & 30.2 \\            
GeM+SiMix\dag                & RS@k & GLDv1~\cite{nas+17} & ours    & 53.1 & 41.8 & 68.4 & 45.3 & 51.0 & 26.4 & 81.2 & 62.4 & 58.7 & 31.1\\            
 \hline
\end{tabular}

\vspace{-1em}
\caption{Performance comparison (mAP\%) on $\mathcal{R}$Oxford and $\mathcal{R}$Paris with 1m distractor images ($\mathcal{R}$1m). Mean performance is reported across all setups or the large-scale setups only. $\ast$ denotes that the FC layer is not part of the training but is added afterward to implement whitening. Batch size is 4096 for all methods; SiMix virtually increases it to 10240. ResNet101 is used as a backbone for all methods.
\label{tab:rop}}
\vspace{-2em}
\end{center}
\end{table*}

\paragraph{Cars196.} Evaluation on a small scale dataset, Cars196~\cite{ksd+13} is presented in the Table \ref{tab:metriclearning}. We train SAP with a batch size of $392$; it provides a performance of $79.5\%$, $86.6\%$, $91.2\%$, and $94.4\%$ and when combined with SiMix a performace of $85.4\%$, $91.0\%$, $94.3\%$ and $96.7\%$ on r@$1$, r@$2$, r@$4$ and r@$8$ respectively. SiMix makes a large difference in performance for both RS@k and SAP~\cite{bxk+20}, primarily because of a smaller batch size ($392$), as constrained by the low number of classes. With SiMix, RS@k reaches the state-of-the-art results on three out of four recall@k values. If the batch size is further increased to $588$ by changing the number of samples per class from $4$ to $6$, then RS@k provides a larger gain with performance $88.3\%$, $93.3\%$, $95.9\%$ and $97.6\%$.

\paragraph{Results with ViT-B.} The results by replacing the ResNet-50~\cite{hzr+16} backbone with a ViT-B~\cite{dbk+21} for SAP~\cite{bxk+20} and the proposed RS@k are also shown in Table \ref{tab:metriclearning}. With an exception of ViT-B/32 on VehicleID and Cars196 datasets, the use of ViT-B backbone leads to better performance for both methods, compared to the ResNet counterpart. It can be clearly seen that RS@k outperforms SAP~\cite{bxk+20} on all datasets. ViT-B/16 when trained with RS@k shows unprecedented performance on all datasets reaching recall@$1$ score of $83.9\%$ on iNaturalist~\cite{vms+18}, $88.0\%$ on SOP~\cite{ohb16}, $96.2\%$ on VehicleID~\cite{ltw+16} (small) and $89.5\%$ on Cars196~\cite{ksd+13}. Note that while ResNet-50 has $24.5$ M parameters and operates with $8.12$ GMac/image, ViT-B has $87.8$ M parameters and operates with $4.36$ and $16.8$ GMac/image for ViT-B/32 and ViT-B/16 respectively.

\paragraph{Concurrent work.} The method of learning intra-batch connections for deep metric learning~\cite{sel21} achieves r@1 of $81.4\%$ on the SOP and $88.1\%$ on Cars196 dataset. The approach for Grouplet embedding learning~\cite{zlx+21} obtains r@1 of $82.0\%$ on SOP and $91.5\%$ on Cars196. The metric mixup approach~\cite{vpa+21} reports the best results of $81.3\%$ r@1 on SOP in combination with ProxyNCA++~\cite{tdt20} and $89.6\%$ on Cars196 which is in combination with MS~\cite{whh+19}.

\paragraph{$\mathcal{R}$Oxford/$\mathcal{R}$Paris.} Table~\ref{tab:rop} summarizes a comparison with AP-based losses in the literature on $\mathcal{R}$Oxford/$\mathcal{R}$Paris with and without distractor images. The comparison is performed with GLDv1 as a training set whose performance is reported for the work of Revaud \etal~\cite{rar+19} in their GitHub page, while the \emph{landmarks-clean dataset} is avoided as all initial images are not publicly available at the moment. During the training performed by us, training images are down-sampled to have a maximum resolution of $1024 \times1024$. 
The inference is performed with multi-resolution descriptors at three scales with up-sampling and down-sampling by a factor of $\sqrt{2}$.
Note that SAP is not evaluated on these datasets in the original work and this experiment is performed by us, which outperforms the previously used AP loss~\cite{hls18}. RS@k, with or without the SiMix, increases the performance by a small margin.

\subsection{Effect of hyper-parameters}
\label{sec:effect_hyperparams}

We study the impact of hyper-parameter on the Cars196 dataset~\cite{ksd+13} since it is the smallest compared to the others and has the lowest training time. 

\begin{figure}
\centering
\raisebox{3pt}{
\input{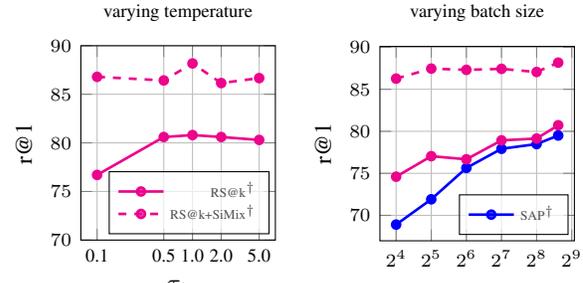}
\begin{tikzpicture}
\begin{axis}[%
	width=0.5\linewidth,
	height=0.5\linewidth,
	xlabel={\small $\tau_{1}$},
	ylabel={\small r@1},
	title={varying temperature},
	legend cell align={left},
	legend pos=south east,
    legend style={cells={anchor=east}, font =\tiny, fill opacity=0.7, row sep=-2.5pt},
   	xtick={0.1,0.5,1.0,2.0,5.0},
   	xticklabels={0.1,0.5,1.0,2.0,5.0},
   	xmode=log,
   	ymin=70,
    grid=both,
]
	\addplot[color=magenta,     solid, mark=*,  mark size=1.5, line width=1.0] table[x=temp, y expr={\thisrow{cars_r1}}] 
	\yfccLambda;
	\addlegendentry{RS@k$^{\dag}$};
	\addplot[color=magenta,     dashed, mark=*,  mark size=1.5, line width=1.0] table[x=temp, y expr={\thisrow{cars_simix_r1}}] \yfccLambda;
	\addlegendentry{RS@k+SiMix$^{\dag}$};
\end{axis}
\end{tikzpicture}
}
\pgfplotstableread{
 		bs_sop	sop_r1     bs_cars     cars_r1      cars_simix_r1   cars_r1_sap
 		16		72.9	    16          74.60       86.27           68.9
 		32		77.3	    32          77.04       87.46           71.91
 		64		79.0	    64          76.67       87.31           75.64
 		128		79.7	    128         78.92       87.43           77.92
 		256		80.5        256         79.14       87.06           78.48
 		512     81.2        392         80.72       88.17           79.50
 		1024    82.0        nan         nan         nan             nan
 		2048    82.5        nan         nan         nan             nan
 		4096    82.8        nan         nan         nan             nan
 		8192    82.4        nan         nan         nan             nan
 	}{\yfccLambda}
\begin{tikzpicture}
\begin{axis}[%
	width=0.5\linewidth,
	height=0.5\linewidth,
	xlabel={\small batch size},
	ylabel={\small r@1},
	title={varying batch size},
	legend cell align={left},
	legend pos=south east,
    legend style={cells={anchor=east}, font =\tiny, fill opacity=0.7, row sep=-2.5pt},
   	xtick={16,32,64,128,256,512,1024,2048,4096},
   	xmode=log,
   	log basis x={2},
    grid=both,
]
	\addplot[color=blue,     solid, mark=*,  mark size=1.5, line width=1.0] table[x=bs_cars, y expr={\thisrow{cars_r1_sap}}]
	\yfccLambda;
	\addlegendentry{SAP$^{\dag}$};
	\addplot[color=magenta,     solid, mark=*,  mark size=1.5, line width=1.0] table[x=bs_cars, y expr={\thisrow{cars_r1}}] 
	\yfccLambda;
	\addplot[color=magenta,     dashed, mark=*,  mark size=1.5, line width=1.0] table[x=bs_cars, y expr={\thisrow{cars_simix_r1}}]
	\yfccLambda;
\end{axis}
\end{tikzpicture}
\vspace{-3.0em}
\caption{The effect of sigmoid temperature $\tau_{1}$ applied on ranks (left) and of batch size (right). Results are shown on Cars196~\cite{ksd+13}.}
\label{fig:ab_tau1_bs}
\vspace{-1.0em}
\end{figure}

\paragraph{Sigmoid temperature $\tau_{1}$ - applied on ranks.}
The effect of the sigmoid temperature $\tau_{1}$ is summarized in Figure \ref{fig:ab_tau1_bs} (left). For both setups of with and without SiMix, $\tau_{1}=1.0$ gives best results while higher and lower values lead to a decline.

\paragraph{Batch size.}
The effect of the varying batch size is shown in Figure \ref{fig:ab_tau1_bs} (right). It demonstrates that large batch size leads to better results. A significant performance boost is observed with the use of SiMix, especially in the small batch size regime, which comes at a small extra computation. A comparison with SAP~\cite{bxk+20} is also shown in this figure. Note that on smaller batch sizes, the proposed RS@k outperforms SAP with a larger margins.
\section{Conclusions}
\label{sec:conclusions}

This work has presented image embedding learning for retrieval using a novel surrogate loss function for the recall@k metric. State-of-the-art results were achieved on a number of standard benchmarks. Training with very large batch size, up to 4k images, has shown to be highly beneficial. The batch size is further increased, in a virtual way, with a newly proposed mixup approach that acts directly on the scalar similarities. This approach offers a boost in performance at a small increase of the computational cost, while its applicability goes beyond the proposed loss. The implementation of the proposed Recall@k Surrogate loss, proposed similarity mixup, along with the training procedure that allows the use of large batch sizes on a single GPU by sidestepping memory constraints, is available at \href{https://github.com/yash0307/RecallatK_surrogate}{https://github.com/yash0307/RecallatK\_surrogate}.

\section*{Acknowledgements}
The authors thank Krist\'{i}na Cinov\'{a} for proofreading. This research was supported by Research Center for Informatics (project CZ.02.1.01/0.0/0.0/16\_019/0000765 funded by OP VVV), by the Grant Agency of the Czech Technical University in Prague, grant No. SGS20/171/OHK3/3T/13, by Project StratDL in the realm of COMET K1 center Software Competence Center Hagenberg, Amazon Research Award, and Junior Star GACR, grant No. GM 21-28830M.
{\small
\bibliographystyle{ieee_fullname}
\bibliography{egbib}
}

\end{document}